\documentclass[10pt,twocolumn,letterpaper]{article}

\usepackage{iccv}
\usepackage{times}
\usepackage{epsfig}
\usepackage{graphicx}
\usepackage{amsmath}
\usepackage{amssymb}
\usepackage{subfigure}
\usepackage{multirow}
\usepackage{bm}

\usepackage{booktabs}
\usepackage{color}
\usepackage[ruled,vlined,algo2e,linesnumbered]{algorithm2e}

\usepackage[pagebackref=true,breaklinks=true,letterpaper=true,colorlinks,bookmarks=false]{hyperref}
\usepackage[capitalize]{cleveref}
\usepackage{colortbl}
\crefname{section}{Sec.}{Secs.}
\Crefname{section}{Section}{Sections}
\Crefname{table}{Table}{Tables}
\crefname{table}{Tab.}{Tabs.}

\iccvfinalcopy 

\def\iccvPaperID{7471} 

\ificcvfinal\fi

\begin{document}

\title{decoupleQ: Towards 2-bit Post-Training Uniform Quantization via decoupling Parameters into Integer and Floating Points}


\author{Yi Guo, Fanliu Kong, Xiaoyang Li, Hui Li, Wei Chen, \\
Xiaogang Tian, Jinping Cai, Yang Zhang, Shouda Liu \\ 
ByteDance\\
{\fontsize{8.6pt}{\baselineskip}\selectfont \texttt \{guoyi.0, kongfanliu.eng, lixiaoyang.x, lihui.sun, chenwei.gavin, tianxiaogang, caijinping.220, zhangyang.elfin, liushouda\}@bytedance.com}
}

\maketitle
\ificcvfinal\thispagestyle{empty}\fi

\begin{abstract}
Quantization emerges as one of the most promising compression technologies for deploying efficient large models for various real time application in recent years. Considering that the storage and IO of weights take up the vast majority of the overhead inside a large model, weight only quantization can lead to large gains. 
However, existing quantization schemes suffer from significant accuracy degradation at very low bits, or require some additional computational overhead when deployed, making it difficult to be applied to large-scale applications in industry.
In this paper, we propose decoupleQ, achieving a substantial increase in model accuracy, especially at very low bits.

decoupleQ abandons the traditional heuristic quantization paradigm and decouples the model parameters into integer and floating-point parts, thus transforming the quantization problem into a traditional mathematical optimization problem with constraints, which is then solved alternatively by off-the-shelf optimization methods.
 Quantization via decoupleQ is linear and uniform, making it hardware-friendlier than non-uniform counterpart, and enabling the idea to be migrated to high-bit quantization to enhance its robustness. 
Our method has achieved well on-line accuracy near fp16/bf16 on the 2-bit quantization of large speech models in ByteDance. 
The code is available at \url{https://github.com/bytedance/decoupleQ}.

\end{abstract}
\section{Introduction}
Serving large models~\cite{zhang2022opt,brown2020language,zhang2023google,bubeck2023sparks} in industry is budget-consuming because of the huge computational, IO and storage cost. Model compression~\cite{guo2021gdp,guo2023rdimkd,krishnamoorthi2018quantizing} has therefore become a necessity to alleviate this pain. Among which, Post-Training Quantization (PTQ)~\cite{nahshan2021loss,frantar2022optq} has gained more and more popularity among researchers and engineers because it does not require heavy GPU-hours training with labeled datasets. In PTQ, weight-only quantization~\cite{lin2023awq, frantar2022optq} plays an important role, since the storage and IO of model weights account for much of the overhead when inference with very large models on low-bandwidth GPUs.

However, previous quantization schemes remain confined within the traditional heuristic quantization paradigm, \eg, how to deal with outliers~\cite{xiao2023smoothquant,wei2023outlier}, how to deal with sensitive channels~\cite{dettmers2022llm}, how to determine the clipping range~\cite{shao2023omniquant}, and so on. These methods have achieved some success, but the quantization at extreme low bit often suffers from significant accuracy degradation, thus failing to meet the launching requirements of industrial practice.
There are also some other options to mitigate the accuracy loss. QuIP~\cite{chee2023quip} pushes the accuracy limits of 2-bit quantization and can achieve performance close to fp16/bf16. However, compared to traditional quantization schemes, its inference imposes an additional burden due to the need to multiply two random orthogonal matrices to de-quant the weights. N2UQ~\cite{liu2022nonuniform} fit the real-value distribution with non-uniform grids then quantize them into equidistant output levels. But it need to train to get the input thresholds. SpQR~\cite{dettmers2023spqr} and SqueezeLLM~\cite{kim2023squeezellm} use mixed-precision quantization or non-uniform scheme to safeguard the important channels, but they need customized hardware support.

In order to alleviate the above pains in industry, 
we proposed decoupleQ, which completely abandons the traditional heuristic quantization paradigm and instead decouples the model parameters into integer and floating point parts, thus transforming the quantization problem into a traditional mathematical constrained optimization problem, which is then solved alternatively by off-the-shelf solution methods. The integer part contains the main weights of the model, and the floating-point part contains scales and zero points induced via quantization. decoulpeQ starts from an abstract objective function and thus does not need to deal with the minutiae of traditional quantization paradigm, such as outlier, salient weights~\cite{lin2023awq}, and so on. Quantization via decoupleQ is linear and uniform, making it hardware-friendlier than non-uniform counterpart, and enabling the idea to be migrated to high-bit quantization to enhance its robustness.

decoupleQ contains two stages: 1. layer-wise minimization, defined in \cref{assumption}, is used to optimize the integer part and the floating-point part; 2. block-wise minimization, defined in \cref{block-min}, is used to further optimize the floating-point part while freezing the integer part\footnote{We define the term ``layer" as a linear transformation, ``block" as a common transformer block containing the multi-head attention, feed forward, and some layer norm.}. 

Layer-wise minimization is widely used in many previous methods~\cite{frantar2022optq,chee2023quip,frantar2022optimal} and works well. For a linear layer, the minimization of the $\ell^2$ loss of the outputs between pre- and post-quantization can be formulated as:
\begin{equation}
    \min_{\widetilde{W}}\|X\widetilde{W}-XW_0\|_2^2
    \label{assumption}
\end{equation}
where $X \in \mathbb{R}^{batch\times d_{in}}$ is the input of this layer, $W_0 \in \mathbb{R}^{d_{in}\times d_{out}}$ is the pre-trained full precision weight, $d_{in}$ and $d_{out}$ are the input and output dimensions respectively. The objective is to find a matrix $\widetilde{W}$ with quantized-then-dequantized elements to minimize \cref{assumption}. 

 Some works~\cite{nagel2020up,hubara2021accurate} started from \cref{assumption} and achieved some success, but they still haven't thought outside the box of traditional quantization. GPTQ series~\cite{frantar2022optq, frantar2022optimal} fake-quantize the first element of $W_0$ and then update the the remaining elements so as to keep \cref{assumption} minimized. This process is then continued element by element until all elements are fake-quantized. However, on the one hand, they do not give any indication of how scale and zero point should be calculated, and on the other hand, the optimization problem formulated when updating the remaining elements is unconstrained (explained in detail later). decoupleQ models \cref{assumption} as a purely mathematical optimization problem, as shown in ~\cref{decoupleQ}. It no longer needs to pay attention to some of the minutiae unique to quantization, such as outliers, clipping threshold, \etc., but abstracts the essence of the problem from a higher level and transforms it into a mathematical constrained optimization problem. 

In the second stage, block-wise minimization is used to further improve the model accuracy:
\begin{equation}
    \min\|\widetilde{\text{Block}(X)}-\text{Block}(X)\|_2^2
    \label{block-min}
\end{equation}
where $\widetilde{\text{Block}(\cdot)}$ is a common transformer block~\cite{vaswani2017attention} with quantized weights. In this stage, we freeze the integer part of the weights, and train the scales and zeros, as well as the parameters in normalization layers.


decoupleQ implements 2-bit uniform quantization and achieves state-of-the-art accuracy in Llama-1/2~\cite{touvron2023llama,touvron2023llama2}. Like traditional uniform quantization, decoupleQ does not incur additional inference burden and only requires a linear transformation to convert the quantized weights into floating point ones.

Our main highlights are summarized as follows:
\begin{itemize}
\item \textbf{New insight:} We abandoned the traditional quantization paradigm, and no longer need to focus on some of the minutiae unique to quantization, but abstracts the essence of the problem from a higher level and transforms it into a constrained optimization problem.

\item \textbf{Extreme low-bit:} decoupleQ achieves 2-bit post-training uniform quantization with performance close to fp16/bf16 for industrial applications in the ASR model in ByteDance.

\item \textbf{Extensibility:} If labeled datasets are available, the idea of decoupleQ can be easily extended to supervised learning to further improve model accuracy, or the adaptation to the downstream sub-tasks.

\end{itemize}

\section{Related Works}

Quantization can be roughly divided into quantization aware training (QAT)~\cite{wei2023outlier,liu2023llm} and Post-Training Quantization (PTQ)~\cite{xiao2023smoothquant,chee2023quip}. In this paper, we focus on weight-only quantization in PTQ, and we will only summarize a few works that are closely related to our work.

PTQ is commonly used for LLM quantization because it does not require a lot of GPU hours of training with labeled datasets. However, in the traditional quantization paradigm, there are many  minutiae specific to quantization that need to be targeted. AdaRound~\cite{nagel2020up} and BRECQ~\cite{li2021brecq} start from the rounding operation and explore whether to round up or down is better. SqQR~\cite{dettmers2023spqr} and OWQ~\cite{lee2023owq} use mixed-precision quantization strategy to protect sensitive parameters, while AWQ~\cite{lin2023awq} opts for scaling up the weights of sensitive channels to reduce the loss of quantization of sensitive channels. OmniQuant~\cite{shao2023omniquant} use gradient decent to optimize for the weight clipping threshold and the rescale factors. In decoupleQ, we abandon patchwork solutions and transform the quantization into a principled traditional optimization problem by decoupling the model parameters into integer and floating-point parts.

GPTQ~\cite{frantar2022optq} is an influential work, and it quantizes the current weights and then updates the remaining weights to minimize the $\ell^2$ loss of the output of the layer between pre- and post-quantization. As we will see later, this update actually approximates much, and GPTQ does not optimize for the scale and zero point reduced by quantization.

QALora~\cite{xu2023qa} also decouples model parameters at a certain level and uses labeled datasets to fine-tune the zero points. decoupleQ takes this idea a step further, optimizing the integer and floating-point parts alternately in the field of PTQ.

\section{Methods}
We introduce the details of decoupleQ in this section. In decoupleQ, we focus on the linear uniform quantization for better hardware efficiency. 
\subsection{Preliminaries}
For a linear layer with input dimension $d_{in}$ and output dimension $d_{out}$, quantization maps the weights with high-precision into discrete level, and the previous scheme can be described as follows:
\begin{equation}
    \widehat{W} = \text{clip}(\lfloor \frac{W_0-z}{s} \rceil, \alpha, \beta)
    \label{quantization}
\end{equation}
\begin{equation}
    \widetilde{W} = \widehat{W}*s+z
    \label{anti-quant}
\end{equation}
where $W_0 \in \mathbb{R}^{d_{in}\times d_{out}}$ is the pre-trained full precision weights, $s$ and $z$ are the scale and zero point (what we call floating-point part above), $\lfloor \cdot \rceil$ is the round-to-nearest function, $\widehat{W} \in \mathbb{R}^{d_{in}\times d_{out}}$ is the quantized integer-point matrix (what we call integer part above), $\widetilde{W}$ is the de-quantized floating-point matrix, $\alpha$ and $\beta$ are the lower and upper bounds of the range of integer representations, respectively. For example,
in 2-bit weight only linear quantization scheme, the value of each entry of $\widehat{W}$ is limited to one of $\{-2, -1, 0, 1\}$, and $\alpha=-2$, $\beta=1$ in this case. To get the values of $\widetilde{W}$, previous methods~\cite{frantar2022optq,frantar2022optimal} show that layer-wise $\ell^2$ loss between the outputs pre- and post-quantization is well related to the model accuracy, \ie, to optimize the following objective function,
\begin{equation}
{\arg\min}_{\widetilde{W}} \|X\widetilde{W}-XW_0\|_2^2 =\text{tr}\{(\widetilde{W}-W_0)^TH(\widetilde{W}-W_0) \}   
\label{optimization}
\end{equation}
where $X \in \mathbb{R}^{batch\times d_{in}}$ is the input of this linear layer, generated by a small set of calibration dataset, and $H=X^TX$.

In the extreme low-bit quantization regime, the model accuracy can be further improved via finer-grained grouping. In this case, the domain of $s$ and $z$ can be expressed as $\mathbb{R}^{d_{out}\times ng}$, where $ng$ is the number of groups, with groupsize $d_{in}/{ng}$. Then, the operations on $s$ and $z$ in Eq.\eqref{quantization} and Eq.\eqref{anti-quant} need to be broadcasted to each group. Finer-grained grouping would impose additional overhead on inference. 
For example, when groupsize=64, it imposes an average overhead of 0.5 bit per element (fp16 for scale $s$ and zero point $z$). The extra overhead is acceptable compared to the model accuracy gain.
\subsection{decoupleQ}
When a model is quantized, only the integer part $\widehat{W}$ and the floating-point part $(s,z)$ in Eq.\eqref{anti-quant} are delivered to the downstream inference engine, and the inference process does not need to know how $\widehat{W}$ and $(s,z)$ are computed at all. That is, if we can find the values of $\widehat{W}$ and $(s,z)$ to minimize \cref{optimization} by other methods, then we don't need to use \cref{quantization} at all. So, we can decouple the model parameters into integer part $\widehat{W}$ and floating point part $(s,z)$, which are then optimized alternatively via off-the-shelf solution methods. decoupleQ views the process of solving for $\widehat{W}$ and $(s,z)$ in Eq.\eqref{anti-quant} as an constrained optimization problem independent of the previous quantization paradigm! We only need to regard Eq.\eqref{anti-quant} as an ordinary affine transformation, in which the value of $s$ can be 0 or even negative. Focusing only on Eq.\eqref{anti-quant} and ignoring Eq.\eqref{quantization} is the core difference between decoupleQ and previous methods.

In per-channel quantization, each column of the weight matrix is optimized independently of each other. For simplicity of notation, we only focus on one column in $\widehat{W}$ later and re-define the notations. Based on Eq.\eqref{optimization}, the optimization problem of decoupleQ in the first stage, layer-wise minimization, can then be formulated as:
\begin{equation}
\begin{aligned}
\min_{w;s,z}& g(w;s,z) \\
\text{s.t.} & \ \forall i=1,2,...,d_{in} \\
& w_i-\beta \le 0   \\
& -w_i+\alpha \le 0 \\
& w_i\in\mathbb{Z} 
\end{aligned}
\label{decoupleQ}
\end{equation}
where the objective function is: 
\begin{equation}
    g(w;s,z)=\frac{1}{2}(w*s+z - b)^TH(w*s+z - b)
    \label{objective function}
\end{equation}
$w\in \mathbb{R}^{d_{in}}$ is one column of $\widehat{W}$, $b\in \mathbb{R}^{d_{in}}$ is the corresponding column of $W_0$, $s\in \mathbb{R}^{ng}$ is the scale and $z\in \mathbb{R}^{ng}$ is the zero point, $ng$ is the number of groups when grouping-quantization. The operations \wrt $(s,z)$, \ie, $*s$ and $+z$, need to be broadcasted to each group.
In this paradigm, we have completely abandoned the traditional framework of quantization and instead transformed quantization into a mathematical optimization problem~\eqref{decoupleQ}, which is solved to achieve the purpose of quantization. $(s,z)$ in problem \eqref{decoupleQ} have lost the meaning of scale and zero point, and are just two simple optimization variables.

Transforming the traditional quantization problem into \cref{decoupleQ} is the soul of decoupleQ! Having completed this shift in thinking, we can then focus on how to solve this optimization problem via off-the-shelf machine learning solution methods. Problem~\eqref{decoupleQ} is a quadratic programming problem with an additional non-convex constraints $w_i\in\mathbb{Z}$. Quadratic programming has been studied for many years and there are now many well-established solution~\cite{murty1988linear,wright2006numerical}. The solving process is not the core contribution of this paper, and we provide one solution in next subsection.

When \cref{decoupleQ} is solved, the model reaches a reasonable accuracy, as shown in the experiment part. The core idea of decoupleQ is to decouple the model weights into the integer part $w$ and the floating-point part $(s,z)$, with the integer part occupying most of the model's expressive power. The extensibility of the idea of decoupleQ is that we can freeze the integer part of the entire model, and use labeled data to train the $(s,z)$ as well as other floating point parameters. The advantage of this is that on the one hand, it can further improve the accuracy of the model, on the other hand, it can fit specific downstream sub-tasks while maintaining the generalization ability of the model. In this paper, we focus on PTQ, thus using only unlabeled datasets to do block-wise minimization, as shown in~\cref{block-min}, to further improve the model accuracy when \cref{decoupleQ} is solved.


\subsection{Optimization via Alternative Iteration}
The problem~\eqref{decoupleQ} is not easy to solve because of the non-convex constraint $w_i\in\mathbb{Z}$. 
After obtaining a good initialization (explained in detail later), we solve for $w$ and $(s,z)$ alternately and iteratively. 
In each round of alternation, the objective function~\eqref{objective function} \wrt $(s,z)$ is an unconstrained quadratic function, thus $(s,z)$ can be readily determined \emph{analytically}: by differentiating the objective function and equating the derivative to zero, followed by solving the resultant linear system of equations. While for $w$, the problem becomes:
\begin{equation}
\begin{aligned}
\min_{w}& g(w;s,z) \\
\text{s.t.} & \ \forall i=1,2,...,d_{in} \\
& w_i-\beta \le 0   \\
& -w_i+\alpha \le 0 \\
& w_i\in\mathbb{Z} 
\end{aligned}
\label{decoupleQ_w0}
\end{equation}
For~\cref{decoupleQ_w0}, one solution is to round-and-clip one element of $w$ to be integer in $[\alpha, \beta]$ and then update the remaining. And then this process is then performed sequentially for all elements. After the $j$-th element has been rounded-and-clipped, the objective for the updating then becomes:
\begin{equation}
\begin{aligned}
\min_{w_i;i>j}& g(w;s,z) \\
\text{s.t.} & \  \forall i=j+1,...,d_{in}  \\
& w_i-\beta \le 0  \\
& -w_i+\alpha \le 0 \\
& w_i\in\mathbb{Z} 
\end{aligned}
\label{optimize remaing0}
\end{equation}

~\cref{optimize remaing0} is also intractable, and we can make two levels of approximation: the first-level approximation is :
\begin{equation}
\begin{aligned}
\min_{w_i;i>j}& g(w;s,z) \\
\text{s.t.} & \  \forall i=j+1,...,d_{in}  \\
& w_i-\beta \le 0  \\
& -w_i+\alpha \le 0
\end{aligned}
\label{level1}
\end{equation}
and the second-level approximation is:
\begin{equation}
\begin{aligned}
\min_{w_i;i>j}& g(w;s,z)
\end{aligned}
\label{level2}
\end{equation}
In the first-level approximation, only the non-convex constraint $w_i \in \mathbb{Z}$ is discarded, while in the second-level approximation, both the non-convex constraint $w_i \in \mathbb{Z}$ and the convex constraint  $w_i \in [\alpha, \beta]$ are  discarded. Intuitively, \cref{level2} is much simpler to solve than \cref{level1}, but solving \cref{level1} will lead to a better convergence of problem~\eqref{decoupleQ} than solving \cref{level2}. GPTQ~\cite{frantar2022optq} provides an efficient analytical solution for \cref{level2}, which we will directly utilize in our experiments. 
( GPTQ updates the remaining elements by considering only the second-level approximation and ignoring the constrain $w_i \in [\alpha, \beta]$ in \cref{level1}, which is what we mentioned in the introduction, that the update of GPTQ is unconstrained.)
As for \cref{level1}, there are many mature solutions in the field of convex optimization, such as active-set method, projected gradient descent, projected coordinate descent and so on~\cite{bubeck2015convex}.  We chose projected gradient descent because its parallelization is much better than the other two methods. In the experimental part, we will compare the final accuracy of the model via between solving \cref{level1} and solving \cref{level2} on small models, while on large models (lager than 7 billion parameters), we have to choose \cref{level2} because the intolerable runtime of solving \cref{level1} many times. The algorithm is shown in \cref{alg1} and \cref{alg2}.

\begin{algorithm2e}[tbp]
 \SetAlgoLined
 \KwIn{ predefined iteration number $N$.}
 \KwResult{$w^*, s^*, z^*$}
 Initialize $t=1, w_0, s_0, z_0$ (initial values); \\
\While{$t\leq N$}{
Freeze $(s_{t-1}, z_{t-1})$, and optimize $g(w;s_{t-1},z_{t-1})$ to obtain an approximate solution $w_t$ via solving \cref{decoupleQ_w0} via \cref{alg2}; \\

Freeze $w_{t}$, and solve the unconstraint quadratic equation $g(w_{t};s,z)$ to obtain an \emph{analytic} solution for $(s_t,z_t)$; \\
$t=t+1$
}
$w^* = w_{N}$; $s^*=s_{N}$; $z^*=z_{N}$
\caption{Alternative Iteration to solve problem~\eqref{decoupleQ}.}
\label{alg1}
\end{algorithm2e}

\begin{algorithm2e}[htbp]
 \SetAlgoLined
 \KwIn{predefined iteration number $K, M$, and the frozen $(s,z)$.}
 \KwResult{$w^*$}
 \uIf {\cref{level1} is used}{
  Ignoring the constraint $w_i \in \mathbb{Z}$ in \cref{decoupleQ_w0}, and train \cref{decoupleQ_w0} with $M$ iterations via projected gradient decent \;}
 Initialize $j=1$; \\
\For{$j=1 \to d_{in}$}{
round and clip the $j$-th element of $w$ to be integer in range [$\alpha$, $\beta$], then keep the first $j$ elements frozen, and update the remaining elements via projected gradient decent to optimize \cref{level1} with $K$ iterations or until converged, or via the method in GPTQ to optimize \cref{level2}.
}
$w^*=w$ 
\caption{Approximate solution of \cref{decoupleQ_w0}}
\label{alg2}
\end{algorithm2e}

\subsection{Initialization of $w$ and $(s,z)$}
Since the values of $w$ are discrete, a good initialization is very important in order to obtain a more accurate solution to the original problem~\eqref{decoupleQ} with a faster convergence. 
Intuitively, the function $g(w;s,z)$ contains the term $w*s$, which means that the scales of the initial values of $w$ and $s$ have to be reasonably distributed. For example, in the extreme case when the initial value of $(s,z)$ have a very large scale, the first iteration will make most of the entries of $w$ strictly 0, which will make the iteration crash. We start by initializing $(s,z)$. We can use grid search to solve the following equation for the initial value of $(s,z)$.
\begin{equation}
\begin{aligned}
\min_{p} & \frac{1}{2}(w*s+z - b)^TH(w*s+z - b) \\
\text{s.t.} & \\
& w= \text{clip}(\lfloor\frac{b-z}{s}\rceil, \alpha, \beta) \\
& s = \frac{p*(b_{max}-b_{bmin})}{\beta-\alpha} \\
& z= p*b_{min}-s*\alpha
\end{aligned}
\label{initialization}
\end{equation}
where $p$ is a single number, may be different for different columns of $W_0$, $b_{min}$ and $b_{max}$ are the minimum and maximum value of $b$ respectively. This step is the same as the previous post-training quantization~\cite{lin2023awq} process. 
Once the grid search is complete, we no longer need to concern ourselves with the $(s,z)$ inside the $\lfloor \cdot \rceil$ function. The point of this step is simply to find an initial value for $(s,z)$ for the optimization problem~\eqref{decoupleQ}.

When solving \cref{decoupleQ_w0} via the first-level approximation (\cref{level1}), before entering the for-loop in \cref{alg2}, we ignore the constraint $w_i \in \mathbb{Z}$ in \cref{decoupleQ_w0} and optimize it via projected gradient decent with $M$ iterations. The purpose of this is to allow \cref{level1} to converge in a small number of iterations, \ie, a small $K$.


\subsection{Block-wise minimization}
After solving problem~\eqref{decoupleQ}, we obtain a solution for the layer-wise minimization stage and a reasonable model accuracy. But minimizing the $\ell^2$ loss at the layer level does not necessarily lead to the minimizing the $\ell^2$ loss at the block level. We found that the model accuracy can be further improved via optimization~\cref{block-min}. BRECQ~\cite{li2021brecq} also shows that block-reconstruction results in a better model accuracy than layer-reconstruction. In this stage, we freeze the integer part $\widehat{W}$ in the whole block and fine-tuning $(s,z)$ and the parameters in norm layer with $J$ epochs.

\section{Experiments}
In this section, we describe in detail the experimental results of our method in comparison with other methods. All the experiments are conducted on a single A100-SXM-80GB. Unless otherwise stated, the default experimental setting is as follows:

\textbf{ResNet:} 10240 images in the training dateloader are used as calibration data, with the standard augmentation in Pytorch official code\footnote{https://github.com/pytorch/examples/blob/master/imagenet/main.py}, and the pretrained full precision checkpoints are from Torchvision~\cite{marcel2010torchvision}. $N=4, M=50$ ($N$ and $M$ is defined in \cref{alg1} and \cref{alg2}). All the convolution layers and fully-connected layers are quantized into W2 without groups.

\textbf{Llama-1/2:} 128 2048-token segments from C4~\cite{raffel2020exploring} are used as calibration data. We choose C4 as calibration dataset instead of WikiText2~\cite{merity2016pointer} to be consistent with GPTQ. If the block-wise minimization is used, we use Adam optimizer~\cite{kingma2014adam} to finetune the $(s,z)$ and the parameters in norm layer with $J=4$ epochs. The learning rate is $1e\textnormal{-}5$, weight decay is $1e\textnormal{-}6$.

\subsection{Private Experiments}
We applied decoupleQ to ByteDance’s Automatic Speech Recognition model(ASR). The input of the model is a speech sequence and some prompt, and the output is the corresponding text. The part of the model that needs to be quantized contains 40 transformer blocks with 13 billion parameters. Word Error Rate (WER) is used as metric to measure the accuracy of the model (less is better). The model is quantized into W2A16g64. In this experiment, we use 3200 pieces of speech containing about 8 millions of tokens as calibration dataset, and train 3 epoch in each block-wise minimization process. The results are shown in ~\cref{tab:private}

\begin{table}[]
    \centering
    \vspace{10pt}
\begin{tabular}{c|cccc}
\hline
        & BF16 & GPTQ & deQ w/o & deQ w/ \\ \hline
WER(\%) & 6.68 & 6.83 & 6.74    & 6.70  \\ \hline
runtime & -    & 10   & 15      & 25    \\ \hline
\end{tabular}
\vspace{10pt}
    \caption{The WER of our ASR model. ``deQ w/o'' means decoupleQ without the block-wise minimization; ``deQ w'' means decoupleQ with both layer-wise minimization and block-wise minimization. The model is quantized into W2A16g64. runtime is measured in hours.}
    \label{tab:private}
\end{table}

\begin{table*}[ht]
\centering
\begin{tabular}{ccccccccc}
\hline
Llama                       &                                            & 1-7B                                  & 1-13B                                 & 1-30B                                 & 1-65B                                 & 2-7B                                  & 2-13B                                  & 2-70B                                 \\ \hline
FP16                        &                                            & 5.68                                  & 5.09                                  & 4.10                                  & 3.53                                  & 5.47                                  & 4.88                                   & 3.31                                  \\ \hline
                            & GPTQ                                       & 2.1e3                                 & 5.5e3                                 & 499.75                                & 55.91                                 & 7.7e3                                 & 2.1e3                                  & 77.95                                 \\
                            & OmniQuant                                  & 15.47                                 & 13.21                                 & 8.71                                  & 7.58                                  & 37.37                                 & 17.21                                  & 7.81                                  \\
                            & \cellcolor[HTML]{9AFF99}\textbf{decoupleQ} & \cellcolor[HTML]{9AFF99}\textbf{9.49} & \cellcolor[HTML]{9AFF99}\textbf{7.86} & \cellcolor[HTML]{9AFF99}\textbf{6.37} & \cellcolor[HTML]{9AFF99}\textbf{5.59} & \cellcolor[HTML]{9AFF99}\textbf{9.74} & \cellcolor[HTML]{9AFF99}\textbf{13.03} & \cellcolor[HTML]{9AFF99}\textbf{5.23} \\
\multirow{-4}{*}{W2A16}     & \cellcolor[HTML]{EFEFEF}runtime            & \cellcolor[HTML]{EFEFEF}2.5           & \cellcolor[HTML]{EFEFEF}4.8           & \cellcolor[HTML]{EFEFEF}12.7          & \cellcolor[HTML]{EFEFEF}27.6          & \cellcolor[HTML]{EFEFEF}2.5           & \cellcolor[HTML]{EFEFEF}4.5            & \cellcolor[HTML]{EFEFEF}33.4          \\ \hline
                            & GPTQ                                       & 44.01                                 & 15.60                                 & 10.92                                 & 9.51                                  & 36.77                                 & 28.14                                  & -                                     \\
                            & OmniQuant                                  & 8.90                                  & 7.34                                  & 6.59                                  & 5.65                                  & 9.62                                  & 7.56                                   & 6.11                                  \\
                            & \cellcolor[HTML]{9AFF99}\textbf{decoupleQ} & \cellcolor[HTML]{9AFF99}\textbf{8.65} & \cellcolor[HTML]{9AFF99}\textbf{7.25} & \cellcolor[HTML]{9AFF99}\textbf{6.04} & \cellcolor[HTML]{9AFF99}\textbf{5.19} & \cellcolor[HTML]{9AFF99}\textbf{8.79} & \cellcolor[HTML]{9AFF99}\textbf{7.44}  & \cellcolor[HTML]{9AFF99}\textbf{4.96} \\
\multirow{-4}{*}{W2A16g128} & \cellcolor[HTML]{EFEFEF}runtime            & \cellcolor[HTML]{EFEFEF}3.7           & \cellcolor[HTML]{EFEFEF}7.7           & \cellcolor[HTML]{EFEFEF}24.3          & \cellcolor[HTML]{EFEFEF}55.0          & \cellcolor[HTML]{EFEFEF}3.7           & \cellcolor[HTML]{EFEFEF}7.9            & \cellcolor[HTML]{EFEFEF}70.6          \\ \hline
                            & GPTQ                                       & 22.10                                 & 10.06                                 & 8.54                                  & 8.31                                  & 20.85                                 & 22.44                                  & -                                     \\
                            & OmniQuant                                  & 8.90                                  & 7.34                                  & 6.59                                  & 5.65                                  & 9.62                                  & 7.56                                   & 6.11                                  \\
                            & \cellcolor[HTML]{9AFF99}\textbf{decoupleQ} & \cellcolor[HTML]{9AFF99}\textbf{8.18} & \cellcolor[HTML]{9AFF99}\textbf{6.96} & \cellcolor[HTML]{9AFF99}\textbf{5.81} & \cellcolor[HTML]{9AFF99}\textbf{5.07} & \cellcolor[HTML]{9AFF99}\textbf{8.41} & \cellcolor[HTML]{9AFF99}\textbf{6.98}  & \cellcolor[HTML]{9AFF99}\textbf{5.34} \\
\multirow{-4}{*}{W2A16g64}  & \cellcolor[HTML]{EFEFEF}runtime            & \cellcolor[HTML]{EFEFEF}4.3           & \cellcolor[HTML]{EFEFEF}8.9           & \cellcolor[HTML]{EFEFEF}27.9          & \cellcolor[HTML]{EFEFEF}64.5          & \cellcolor[HTML]{EFEFEF}4.4           & \cellcolor[HTML]{EFEFEF}9.0            & \cellcolor[HTML]{EFEFEF}98.2          \\ \hline
                            & GPTQ                                       & 8.06                                  & 6.76                                  & 5.84                                  & 5.06                                  & 8.37                                  & 6.44                                   & 4.82                                  \\
                            & AWQ                                        & 11.88                                 & 7.45                                  & 10.07                                 & 5.21                                  & 24.00                                 & 10.45                                  & -                                     \\
                            & OmniQuant                                  & 6.49                                  & 5.68                                  & 4.74                                  & \textbf{4.04}                                  & 6.58                                  & \textbf{5.58}                          & 3.92                                  \\
\multirow{-4}{*}{W3A16}     & \cellcolor[HTML]{9AFF99}\textbf{decoupleQ} & \cellcolor[HTML]{9AFF99}\textbf{6.38} & \cellcolor[HTML]{9AFF99}\textbf{5.60} & \cellcolor[HTML]{9AFF99}\textbf{4.67} & \cellcolor[HTML]{9AFF99}6.05             & \cellcolor[HTML]{9AFF99}\textbf{6.22} & \cellcolor[HTML]{9AFF99}5.72           & \cellcolor[HTML]{9AFF99}\textbf{3.84} \\ \hline
                            & GPTQ                                       & 6.13                                  & 5.40                                  & 4.48                                  & 3.83                                  & 5.83                                  & 5.13                                   & 3.58                                  \\
                            & AWQ                                        & 6.08                                  & 5.34                                  & 4.39                                  & 3.76                                  & 6.15                                  & 5.12                                   & -                                     \\
                            & OmniQuant                                  & 5.86                                  & 5.21                                  & 4.25                                  & 3.71                                  & 5.74                                  & \textbf{5.02}                          & 3.47                                  \\
\multirow{-4}{*}{W4A16}     & \cellcolor[HTML]{9AFF99}\textbf{decoupleQ}          & \cellcolor[HTML]{9AFF99}\textbf{5.85} & \cellcolor[HTML]{9AFF99}\textbf{5.21} & \cellcolor[HTML]{9AFF99}\textbf{4.24} & \cellcolor[HTML]{9AFF99}\textbf{3.67} & \cellcolor[HTML]{9AFF99}\textbf{5.70} & \cellcolor[HTML]{9AFF99}5.06           & \cellcolor[HTML]{9AFF99}\textbf{3.45} \\ \hline
\end{tabular}
\vspace{10pt}
\caption{The results of PPL of wikitext-2 on Llama-1/2. We also report the runtime (measured in hours) for the W2 quantization via decoupleQ in the gray background row. The results other than decoupleQ are copied from OmniQuant~\cite{shao2023omniquant}. All the results of decoupleQ use the approximation~\cref{level2}. The PPL on Llama-2-13B-W2A16 is higher than that on Llama-2-7B-W2A16: This is strange, and as of press time, we still don’t know the reason.}
\label{ppl}
\end{table*}

\subsection{Public Comparison}
As a first comparison, we compare decoupleQ with other methods on ImageNet~\cite{deng2009imagenet} on ResNet~\cite{he2016deep}, which are standard benchmarks and are easy to implement. Most importantly, its Top-1 is a strong indicator of model accuracy.
\cref{tab:public_resnet} shows the results of decoupleQ and others. The results other than decoupleQ are copied from GPTQ~\cite{frantar2022optq}. \cref{tab:resnet_w2} shows the results of W2 quantization via decoupleQ.

~\cref{ppl} shows the results on Llama. In this experiment, we have to choose \cref{level2} because the intolerable runtime of solving \cref{level1} many times. For a fair comparison, the calibration dataset contains 128 samples, although a larger calibration dataset will result in stronger results. we can see that decoupleQ outperforms others in all settings, although we use a weaker approximation, \cref{level2} instead of \cref{level1}, to save time. As for the hype-parameters, we choose $\{N=4,J=4\}$.

\subsection{Ablation studies}
\subsubsection{the two approximations}
The soul of decoupleQ is \cref{decoupleQ}, but when solving \cref{decoupleQ}, we have to take some approximations, \cref{level1} or \cref{level2}. Obviously, solving \cref{level1} will be much more time consuming than solving \cref{level2}. But if solving \cref{level1} yields better results, the time cost may be worth it. We first evaluate these two approximations from the perspective of model accuracy. In practice, we don't have to wait for \cref{level1} to fully converge when we solve it via projected gradient decent, and only need to iterate some steps to get a sub-optimal solution.
In \cref{alg2}, the for-loop takes up the majority of the runtime. So, we first study the influence of the number of iterations $K$ (defined in the for-loop) on the final accuracy of the model. 

\begin{figure}[htbp]
    \centering
    \includegraphics[width=0.9\columnwidth]{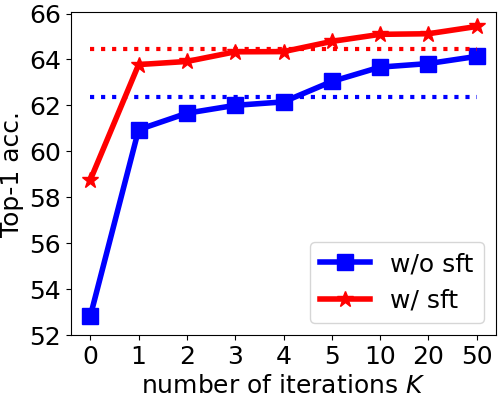}
    \caption{The solid lines represent the top-1 accuracy of ResNet-18 on ImageNet \wrt the number of iterations $K$ when using approximation~\cref{level1}; while the dashed lines are for the approximation~\cref{level2}. The \textcolor[rgb]{0,0,1}{blue} line represents quantization via decoupleQ, with only the layer-wise minimization used. The \textcolor[rgb]{1,0,0}{red} line represents the addition of one-epoch sft to the \textcolor[rgb]{0,0,1}{blue} line.}
    \label{top1_with_K}
\end{figure}

\cref{top1_with_K} shows the Top-1 accuracy of ResNet-18 on ImageNet \wrt the number of iterations $K$. First of all, in the \textcolor[rgb]{0,0,1}{blue} line, we use only the layer-wise minimization of decooupleQ to quantize the model. After the quantization is finished, in the \textcolor[rgb]{1,0,0}{red} line, we use the labelled dataset with the common 1.2 millions images to fine-tune $(s,z)$ and parameters in norm layers, with the integer part being frozen. In this step, we use SGD optimizer with learning rate $1e$-$6$, weight decaying rate $1e$-$4$ to train for only one epoch.
\cref{top1_with_K} clearly indicates the following conclusions: 1. As the number of iterations $K$ increases, the model accuracy increases almost monotonically; 2. When $K>4$, approximation via \cref{level1} is better than via \cref{level2}. This is to be expected, since \cref{level2} drops the constraint $\alpha \le w_i \le \beta$, leading to a looser approximation; 3. By the supervised fine-tuning (sft), the model accuracy is further improved. The same experimental phenomenon also occurs on the ResNet-50 model, which we do not show here.

\begin{figure}
    \centering
    \includegraphics[width=0.9\columnwidth,trim=0 0 30 20,clip]{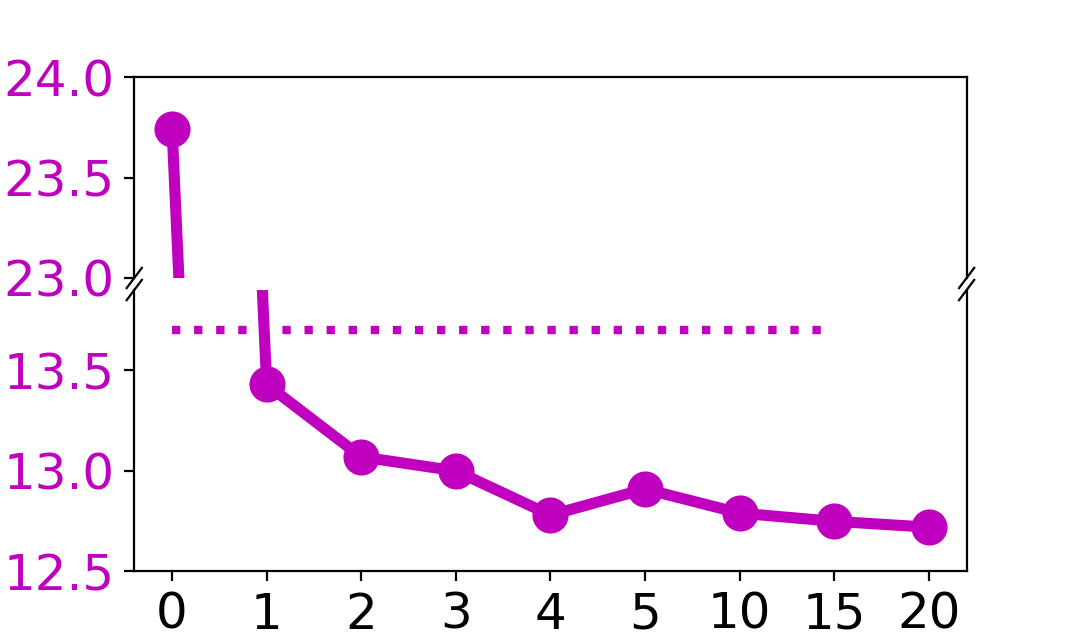}
    \caption{The solid line represents the PPL of Llama-7B on WikiText2 \wrt the number of iterations $K$ when using approximation~\cref{level1}; while the dashed line is for the approximation~\cref{level2}. The horizontal axis represents $K$, and the vertical axis represents PPL. The model is quantized into W2A16, and block-wise minimization is not used in this experiment. It shows that, when $K>1$, solving approximation~\cref{level1} yields better model accuracy than approximation~\cref{level2}.}
    \label{ppl_with_K}
\end{figure}

In the experiment shown in ~\cref{ppl_with_K}, we randomly select 512 2048-token segments from C4~\cite{raffel2020exploring}. We chose 512 segments here instead of the common 128 in order to reduce the effect of overfitting and thus compare the two approximations more objectively. In this experiment, we take $N=2$, and quantize Llama-7B into W2A16 without groups, and only the layer-wise minimization is used to exclude the interference of other factors. The PPL decrease almost monotonically as the number of iterations $K$ increases.

\begin{figure}
    \centering
    \includegraphics[width=0.9\columnwidth]{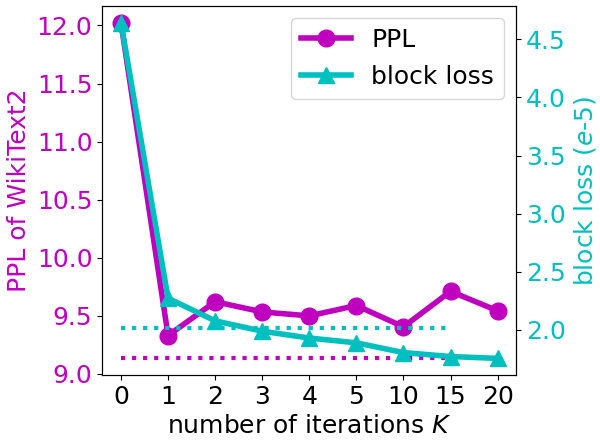}
    \caption{The PPL of Llama-7B on WikiText2 and the loss of the first block between pre-and post-quantization \wrt the number of iterations $K$ when using approximation~\cref{level1}. The dashed line is for the approximation~\cref{level2}. The model is quantized into W2A16, and both the layer-wise minimization and block-wise minimization are used. 
    The model's best PPL is where $K=1$, and then fluctuates within a range as $K$ increases. But all PPLs are inferior to when the approximation ~\cref{level2} is used. The loss, defined in \cref{block-min}, of the first block between pre-and post quantization is plotted on the right vertical axis. As $K$ increases, the loss decreases strictly monotonically, and when $K > 2$, the loss falls below the case when the approximation~\cref{level2} is used.}
    \label{loss_and_ppl}
\end{figure}

However, when block-wise minimization is introduced in addition to the experiment in~\cref{ppl_with_K}, the situation becomes a little more elusive. The results are shown in \cref{loss_and_ppl}. 
The model's best PPL is where $K=1$, and then fluctuates within a range as $K$ continues to increase. But all PPLs are inferior to when the second-level approximation (\cref{level2}) is used. We also plot the loss, defined in \cref{block-min}, of the first block between pre-and post quantization on the right vertical axis. As $K$ increases, the loss decreases strictly monotonically, and when $K > 2$, the loss falls below the case when the approximation~\cref{level2} is used. This suggests that the correlation between PPL and loss is perhaps weak, and we will investigate this in the future.

\begin{figure}
    \centering
    \includegraphics[width=0.9\columnwidth]{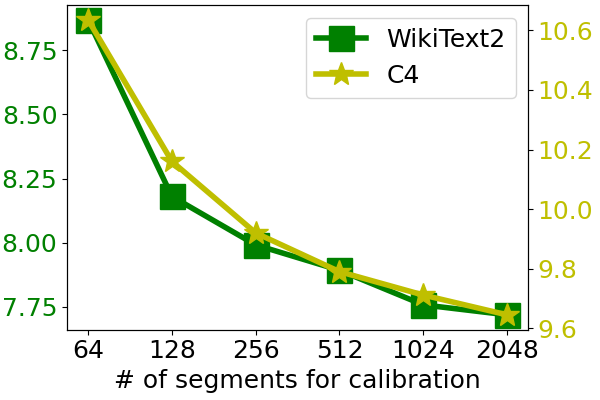}
    \caption{The perplexity of Llama-7B on WikiText2 and C4 dataset \wrt the number of segments as calibration datasets. The model is quantized into W2A16g64.}
    \label{ppl_with_dataset}
\end{figure}

\subsubsection{the size of calibration dataset}
The solution of \cref{decoupleQ} is dependent on $H$ and thus on the size of the calibration dataset, as does \cref{block-min}. Smaller calibration datasets can easily lead to overfitting. \cref{ppl_with_dataset} shows the relationship between dataset size and PPL. In this experiment, the calibration dataset is randomly sampled from C4 and the model (Llama-7B) is quantized into W2A16g64. We use the second-level approximation (\cref{level2}) to save time, and \{$N=4,J=4$\}. It is obvious that as the size of the dataset increases, the model accuracy increases monotonically. For runtime reference, when the number of segments is 128/2048, the experiment took 4.3/19.5 hours.

\subsubsection{the necessity of block-wise minimization}
\cref{tab:without second stage} shows that block-wise minimization,~\cref{block-min}, can further improve the model accuracy. This is obvious because minimizing the $\ell^2$ loss at each layer does not necessarily result in the minimizing the $\ell^2$ loss for this block. In this experiment, we choose $N=4$ and the approximation ~\cref{level2} for the layer-wise minimization, and $J=4$ if block-wise minimization is used.

\begin{table}[t]
    \centering
\begin{tabular}{cccccc}
\hline
Llama & 1-7B   & 1-13B & 1-30B & 2-7B   & 2-13B  \\ \hline
w/o   & 13.66 & 9.68 & 7.35     & 14.66 & 12.93 \\
w     & 9.49   & 7.86  & 6.37  & 9.74   & 13.03  \\ \hline
\end{tabular}
\vspace{10pt}
    \caption{The perplexity of Llama on WikiText2 with and without the block-wise minimization. All the models are quantized into W2A16.}
    \label{tab:without second stage}
\end{table}

\begin{table}[]
    \centering
\begin{tabular}{c|cc|cc}
\hline
\multirow{2}{*}{method} & \multicolumn{2}{c|}{res18-69.76\%} & \multicolumn{2}{c}{res50-76.13\%} \\
                        & 3bit             & 4bit            & 3bit            & 4bit            \\ \hline
GPTQ                    & 67.88            & 69.37           & 74.87           & 75.71           \\
OBQ                     & 68.69            & 69.56           & 75.24           & 75.72           \\
BRECQ                   & 68.47            & 69.37           & 75.32           & 75.88           \\
decoupleQ               & 68.65            & 69.58           & 75.24           & 76.00           \\
decoupleQ*              & 68.94            & 69.71           & 75.61           & 75.97           \\ \hline
\end{tabular}
\vspace{10pt}
    \caption{Comparison of decoupleQ with other methods. In decoupleQ*, we train the $(s,z)$ and parameters in norm layer for one epoch, using the regular labeled dataset containing 1.2 million images.}
    \label{tab:public_resnet}
\end{table}

\begin{table}[]
    \centering
\begin{tabular}{c|cc|cc}
\hline
       & \multicolumn{2}{c|}{decoupleQ} & \multicolumn{2}{c}{decoupleQ*} \\ \hline
ResNet & res18          & res50         & res18          & res50         \\
Top-1  & 64.13          & 71.54         & 65.44          & 72.72         \\ \hline
\end{tabular}
\vspace{10pt}
    \caption{The results of W2 quantization via decoupleQ. In decoupleQ*, we train the $(s,z)$ and parameters in norm layer for one epoch, using the regular labeled dataset containing 1.2 million images.}
    \label{tab:resnet_w2}
\end{table}

\section{Conclusion and Discussion}
deocupleQ decouples the model parameters into the integer part and a floating point part, and then optimizes them alternately. This optimization process contains two stages. In the layer-wise minimization, we transform the quantization problem into the purely mathematical constrained optimization problem~\cref{decoupleQ}; while in the block-wise minimization, we freeze the integer part and then finetune the floating point part.

The risk of decoupleQ comes from two aspects. On the one hand, how much the minimization of the $\ell^2$ loss of the layer's or block's output correlates with the accuracy of the model; on the other hand, decoupleQ is prone to overfitting the calibration dataset. 

For the first risk, we find experimentally that the correlation between Top-1 and the loss is strong in the Imagenet classification task; however, the correlation between PPL and the loss is slightly weaker in LLM. This could be mainly because of an inherent bias between the loss and the accuracy of the model, or because PPL is not a good indicator of the accuracy of LLM, or for other reasons. For the second risk, when $H$ in ~\cref{objective function} is an underdetermined matrix, the risk of overfitting rises sharply. In this case, the possibility of $H$ being underdetermined can be reduced either by enhancing the diagonal element values of $H$ or by increasing the amount of calibration data. In our practice, we found that the accuracy of quantization models can rise monotonically with the increase of the size of the calibration dataset in any situations, but the runtime of quantization rise as well.

The idea of decoupleQ is helpful for the adaptation of large model to downstream sub-task. We can quantize a large foundation model via decoupleQ, then freeze the integer part of the model, and finetune the floating-point part with labeled dataset from downstream sub-task. \cref{tab:resnet_w2} and \cref{top1_with_K} show that the model accuracy can be further improved by end-to-end supervised learning.

{\small
\bibliographystyle{ieee_fullname}
\bibliography{egbib}
}

\end{document}